\documentclass[preprint]{elsarticle}

\usepackage{hyperref}
\usepackage[ruled,vlined]{algorithm2e}
\SetArgSty{textnormal}

\usepackage{siunitx}
\usepackage{amsthm}
\usepackage{amssymb}
\usepackage{amsmath}

\usepackage{lscape}

\theoremstyle{definition}

\makeatletter
\def\ps@pprintTitle{%
	\let\@oddhead\@empty
	\let\@evenhead\@empty
	\let\@oddfoot\@empty
	\let\@evenfoot\@oddfoot
}
\makeatother

\begin{document}

\begin{frontmatter}

\title{Hub-aware Random Walk Graph Embedding Methods for Classification}

\author[ouraddress]{Aleksandar Tomčić\corref{mycorrespondingauthor}}
\cortext[mycorrespondingauthor]{Corresponding author}
\ead{aleksandart@dmi.uns.ac.rs}

\author[ouraddress]{Miloš Savić}
\ead{svc@dmi.uns.ac.rs}

\author[ouraddress]{Milo\v{s} Radovanovi\'c}
\ead{radacha@dmi.uns.ac.rs}

\address[ouraddress]{University of Novi Sad, Faculty of Sciences, Department of Mathematics and Informatics\\Trg Dositeja Obradovi\'ca 4, 21000 Novi Sad, Serbia}

\begin{abstract}
In the last two decades we are witnessing a huge increase of valuable big data structured in the form of graphs or networks. 
To apply traditional machine learning and data analytic techniques to such data it is necessary to transform graphs into
vector-based representations that preserve the most essential structural properties of graphs. For this purpose, a large number
of graph embedding methods have been proposed in the literature. Most of them produce general-purpose embeddings suitable for a variety
of applications such as node clustering, node classification, graph visualisation and link prediction.
In this paper, we propose two novel graph embedding algorithms based on random walks that are specifically designed 
for the node classification problem. Random walk sampling strategies of the proposed algorithms have been designed to pay special 
attention to hubs -- high-degree nodes that have the most critical role for the overall connectedness in large-scale graphs.
The proposed methods are experimentally evaluated by analyzing the classification performance of three classification algorithms
trained on embeddings of real-world networks. The obtained results indicate that our methods considerably improve the predictive
power of examined classifiers compared to currently the most popular random walk method for generating general-purpose 
graph embeddings (node2vec).

\end{abstract}

\begin{keyword}
graphs, networks, hubs, random walks, graph embeddings, node2vec, node classification
\end{keyword}

\end{frontmatter}


\section{Introduction}

Many complex systems can be naturally represented by graphs or networks showing interactions among constituent elements~\cite{Newman2010}. Typical examples include engineered systems (e.g., Internet, power grids, transportation systems, IoT systems), biological systems (e.g., metabolic networks, protein interactions, genetic regulatory networks, brain networks, food webs), knowledge and information systems (e.g., WWW, semantic web taxonomies, citations among scientific papers, recommender networks, linguistic networks) and social systems (e.g., collaboration in science, industry and other forms of social organizations; interactions at online social networks). In the past two decades, motivated by the seminal papers of Watts and Strogatz~\cite{watts_collective_1998} and Barabasi and Albert~\cite{Barabasi99emergenceScaling}, researchers have analyzed structure, function and evolution of many large-scale complex networks from various domains indicating common properties such as heavy-tailed distributions of node connectedness and centrality metrics, assortative and disassortative mixing patterns, dense ego-networks, core-periphery and community structures, evolution governed by some form of preferential attachment, small-world properties, evolutionary densification and shrinking diameters, etc. Those empirical studies initiated a new interdisciplinary research field known as network science whose focus is on exploratory data analysis methods and predictive modeling tools for graph-structured data. 

For many networks labels indicating categories or classes could be assigned to nodes~\cite{Savic2018,NewmanClauset2016}. For example, in paper citation networks node labels indicate research fields, in scientific co-authorship networks node labels correspond to the institutional affiliation of researchers, in the WWW network node labels could denote topics of WWW pages, etc. In the case of large-scale networks, due to a large number of nodes, label assignments are typically partially given, which means that we have both labeled and unlabeled nodes. The task of inferring labels for unlabelled nodes relying on labelled nodes and the network structure is known as the node classification problem. This problem, together with the problem of identifying community structures and the problem of predicting future links, is one of the most important algorithmic problems in the network science.

The problem of node classification can be approached in several ways. The three most dominant approaches in the literature are collective classification~\cite{senaimag08}, node classification based on graph embeddings~\cite{Xu2021SIAM,Cai2018,GOYAL2018} and graph neural networks~\cite{WU2021GNN,ZHOU202057}. A collective classification algorithm consists of three components: a local classifier, a relational classifier and a collective inference scheme. The local classifier estimates initial labels (or the distribution of label probabilities) of unlabeled nodes. The role of the relational classifier is to assign the label (or the distribution of label probabilities) to an individual unlabeled node considering the known labels in its neighborhood. The iterative classification scheme defines how the relational classifier is iteratively applied to obtain labels of all unlabeled nodes. 

The problem of node classification can be reduced to the problem of traditional classification by machine learning algorithms designed for tabular data. This reduction is enabled by graph embedding algorithms. Graph embedding algorithms learn numerical representation of nodes typically by preserving graph-based distances in an Euclidean space of an arbitrarily given dimension. Then, some traditional classification learning algorithm is applied to the table composed of feature vectors of all labeled nodes to obtain the classification model able to infer labels of unlabeled nodes. 
   
Graph neural networks (GNNs) are node classification methods suitable for attributed networks (networks in which each node is represented by a local feature vector independent of the network structure). They typically employ a message passing scheme in which each node aggregates feature vectors of its neighbors to derive the feature vector for the next iteration. After $k$ message passing iterations, a node is represented by its transformed feature vectors capturing the structural information of all nodes at the shortest-path distance equal to $k$. The weights controlling feature vector aggregation can be learned to minimize a loss function considering labeled nodes (e.g., categorical cross-entropy) in order to obtain the classification model for unlabelled nodes.    

The most fundamental metric reflecting the importance of nodes is node degree -- the number of links attached to a node or the number its nearest neighbors. Real world large-scale graphs have power-law~\cite{RekaBarabasi2002} or some other long-tailed distribution~\cite{Clauset2009} of node degrees. Such distributions imply the presence of so-called hubs: nodes with an exceptionally large degree, much larger than the average degree. Although hubs have a vital role for the overall connectedness of complex networks (the so-called ``robust yet fragile'' property stating that the removal of a small fraction of hubs leads to an extremely fragmented network without a giant connected component~\cite{albert2000error}), they have not been seriously considered when designing algorithms for the node classification problem. Additionally, in labeled graphs we can distinguish ``good'' and ``bad'' hubs~\cite{radovanovic2010hubs}: good hub tends to be dominantly surrounded by nodes having the same label, whereas the label of a ``bad'' hub is different than the most frequent label in its neighborhood. In this paper we propose two novel graph embedding methods tailored for the node classification problem that are based on biased random walking strategies taking into account node labels and hubness properties. The proposed methods are experimentally analyzed by evaluating traditional classifiers (support vector machines, na\"ive Bayes and random forests) on obtained graph embeddings. As the baseline method for comparison we take node2vec~\cite{nodetovec2016}, which is the state-of-the-art graph embedding algorithm based on the most general biased random walking strategy. 

The rest of the paper is structured as follows. Section~\ref{rw} gives an overview of existing approaches for the node classification problem. The next section (Section~\ref{mc}) outlines the motivation for this work and its contributions. The proposed hub-based graph embedding methods for node classification are described in Section~\ref{brwm}. The obtained experimental results are presented in Section~\ref{experiments_results}. The last section concludes the paper and gives directions for possible future work.

\section{Related Work}
\label{rw}

\subsection{Collective Classification}

Traditional classification algorithms assume that the data instances in a training dataset are independent of each other. However, in networks we have connected data instances (nodes connected by links). Moreover, the presence of homophily (i.e., the tendency that similar nodes tend to be connected among themselves) in many complex networks implies that there is a strong correlation between labels of connected nodes~\cite{senaimag08}. To effectively take this correlation into account, collective classification methods make a first-order Markov dependency assumption stating that the label of a node depends on the labels present in its neighborhood~\cite{Macskassy2007}.

The most general framework for collective classification was proposed by Macskassy and Provost~\cite{Macskassy2007}. They have been formalized collective classification as a composition of three models:
\begin{enumerate}
 \item Non-relational or local model. This model set initial node labels in isolation using only local node attributes. Such models can be trained by traditional machine learning algorithms. If nodes do not have any additional attributes except single labels, the local model either assumes an equal probability for each label or assigns label probabilities proportionally to label frequencies.
 \item Relational model. The main objective of the relational model is to infer the label or the distribution of label probabilities for a given node considering the known labels in its neighborhood. The most commonly used relational models are the weighted vote relational neighbour classifier, the class-distribution relational neighbour classifier and network-only Bayes classifier~\cite{Macskassy2007}.
 \item Collective inference model. This model controls how the relational model is applied to simultaneously infer labels of all unlabeled nodes. Typical collective inference models are Gibbs sampling, relaxation labeling, iterative classification and loopy belief propagation~\cite{Macskassy2007,senaimag08}.
\end{enumerate}

It has been demonstrated that the injection of unobserved (non-existent) links can improve the performance of collective classification. Galagher et al.~\cite{Gallagher2008} designed a scheme in which ``ghost'' edges are added for unlabeled nodes entirely surrounded with other unlabeled nodes. Such nodes are by ghost edges artificially connected to nearest labeled nodes (nearest by the shortest path distance). The collective inference algorithm proposed in~\cite{Bilgic2007} combines collective inference with link prediction, i.e., links indicated by a link prediction algorithm are injected into a network prior to collective inference.

Collective inference methods can be also generalized for the multi-class setting in which nodes could have multiple labels~\cite{Ghamrawi05,KongSY11}. Cautious inference~\cite{McDowell2009Cautious}, active inference~\cite{Bilgic2010Active} and label regularization~\cite{McDowell2012} were also considered as additional corrective mechanisms increasing robustness of collective classification.

\subsection{Classification based on Graph Embeddings}

Graph embedding algorithms enable the application of traditional machine learning techniques designed for tabular data on graphs. The main idea of graph embedding algorithms is to learn a latent node representation in an Euclidean space of an arbitrary dimension $n$. This means that each node is represented as an $n$-dimensional real-valued vector. The representation learning process is based on the following principle: nodes close or similar in the graph (by a graph-based similarity or distance function) should be also close in the embedding (by a distance defined in the Euclidean space). The embedding produced from a graph together with known node labels can be treated as an ordinary data table for traditional machine learning algorithms training classifiers (e.g., support vector machines, na\"ive Bayes, random forests, etc.). 

Most of the graph embedding algorithms can be, roughly speaking, divided into the following 3 broad categories~\cite{GOYAL2018}:
\begin{enumerate}
 \item methods based on matrix factorization,
 \item methods based on random walks, and
 \item methods based on deep learning techniques.
\end{enumerate}
Methods from the all three indicated categories could be abstracted into an encoder-decoder schema: the encoder transforms a node into its embedding vector, whereas the decoder reconstructs the neighborhood of the node from the embedding vector~\cite{Hamilton2017}. 

Let $A$ denote the adjacency matrix of a graph, a matrix derived from the adjacency matrix or some other matrix reflecting similarity of all node pairs. Lower-dimensional representations of $A$ (node embedding vectors) can be learned by matrix factorization methods that decompose $A$ into a product of two or more matrices. The main property of matrix factorization methods is that the product decomposition is obtained by optimizing a stated loss function. For example, in the Laplacian eigenmap method~\cite{Belkin2001}, $A$ is the graph Laplacian matrix and its factorization is obtained by optimizing a loss function taking into account the L2 distance of node embedding vectors for non-zero values in $A$. A comprehensive overview of matrix factorization methods for producing graph embeddings can be found in~\cite{Cai2018}.

Graph embedding methods based on random walks reduce the problem of generating graph embeddings to the problem of generating text embeddings. More specifically, a certain number of random walks is sampled for each node. The sampled random walks are then treated as ordinary sentences in which words are node identifiers. Since random walks are transitions from a node to one of its randomly selected neighbors, nodes close in the graph will be also close in the text. Random walk graph embedding algorithms typically use word2vec~\cite{Mikolov2013Ext} to form embeddings from sampled random walks. 
DeepWalk~\cite{deepwalk2014} is the first proposed graph embedding algorithm based on random walks. Its main characteristic is that it samples unbiased random walks: each neighbor of the node at which the random walk is currently residing has an equal probability to be selected as the next node. Node2vec is the most general algorithm of this kind~\cite{nodetovec2016}. It has two hyper-parameters for controlling biased random walk sampling that interpolates between BFS and DFS random walking strategies. Savi\'{c} et al.~\cite{2021_NCLID_SISAP} proposed two node2vec extensions that personalize node2vec hyper-parameters per nodes and edges according to a measure based on the notion of local intrinsic dimensionality. More specific random sampling techniques could also be designed to preserve some higher-order graph properties such as structural roles~\cite{struc2vec}.

Graph embedding methods based on deep learning techniques produce graph embeddings by forming autoencoders. Autoencoders are neural networks trained to reproduce input values at the output layer through an architecture composed of an encoder and a decoder. The encoder is a sequence of layers where each next layer contains less neurons than the previous layer. The last layer in the encoder is the middle layer encoding the latent lower-dimensional representation of the training dataset. The decoder starts after the middle layer and its structure is inversed to the encoder. Autoencoders for graphs can by formed either to reproduce adjacency matrices~\cite{Wang2016} or matrices containing values of node similarity metrics~\cite{Cao2016}.

\subsection{Classification based on Graph Neural Networks}

Node classification models can be obtained by training graph neural networks (GNNs). The main feature of GNNs is that they use message passing in which vector representations of nodes are exchanged among neighbors in a graph and updated using deep learning techniques~\cite{Hamilton2020}. In each message passing iteration, a node aggregates vector representations of its neighbors and then updates its own representation. As the message passing procedure advances the node aggregates more information from more distant nodes in the graph.
The functions for aggregating and updating vector representations are arbitrary differentiable functions, so $K$ iterations of the message passing procedure can be viewed as a neural network of $K$ layers. The last iteration produces the final vector representations of nodes. Therefore, a GNN can be trained to minimize a loss function defined with respect to the known labels in the graph in order to obtain the node classification model. 

Graph convolutional networks (GCNs) are the most dominant graph neural network models~\cite{kipf2017semisupervised}. Mathematically speaking, a GCN is a chain of matrix multiplications involving a normalized adjacency matrix, node feature matrix (labels and other node attributes) and trainable weight matrix, where matrix multiplications at the end of each iteration are followed by an activation function (e.g., the ReLU function). Weight matrices (one per layer/iteration) are trained to minimize the cross-entropy error over all labeled nodes. It was demonstrated that the performance of GCNs could be improved by importance sampling~\cite{chen2018fastgcn}, localized spectral filtering~\cite{Defferrard2016} and attention mechanisms~\cite{velickovic2018graph}. A comprehensive review of GCNs can be found in~\cite{Zhang2019GCN}.

Hamilton et al. introduced GraphSAGE -- a framework for inductive representation learning on large graphs~\cite{Hamilton2017GraphSAGE}. The main idea of GraphSAGE is not to learn vector representations of nodes, but a function that performs sampling and aggregation of features from a node's local neighborhood. This work inspired many researchers to investigate generalized neighborhood aggregation mechanisms in GNNs, including skip connections, the Janossy pooling, gated updates inspired by recurrent neural networks and jumping knowledge connections~\cite{Hamilton2020}.

\section{Motivation and Contributions}
\label{mc}

Collective classification methods and graph neural networks can be considered as ``direct'' methods for the node classification problem. In general, graph neural networks are more powerful than collective classification algorithms due to more sophisticated neighborhood aggregation mechanisms, especially in the case of nodes enriched with discriminative attributes (attributes that correlate with class labels). However, graph neural networks demand more time to train and their performance strongly depends on a large space of hyper-parameters whose tuning is also non-trivial task requiring considerable time.

In contrast to collective classification and graph neural networks, classification based graph embeddings is an ``indirect'' approach to infer unknown labels providing a valuable flexibility regarding the final classification model: any traditional classification model designed for tabular data, including also modern deep learning models, can be trained on the top of a graph embedding. Since good graph embeddings preserve nodes' neighborhoods, classification models trained on them are able to achieve at least the same level of performance as collective classification methods. Moreover, graph embeddings based on random walks capture broader neighborhoods of nodes since random walks are not strictly restricted to ego networks (direct neighbors of a node) making their performance comparable to graph neural networks in case when nodes have sparse attributes or do not have attributes at all. On the other hand, classification based on graph embeddings could be significantly more time efficient than tuning and training graph neural networks. Consequently, it can be said that this approach provides a good trade-off between classification performance and training time.

The focus of this paper is on node classification based on graph embeddings obtained via random walks. To the best of our knowledge, the previous research works have not considered random walk graph embedding algorithms specifically tailored for the node classification problem, but they dealt with methods for obtaining general purpose graph embeddings that can be used in a variety of applications (besides node classification, node clustering and link prediction are the most considered applications of graph embeddings). This means that labels of nodes have to be explicitly taken into account when performing random walks. In this paper we propose two novel graph embedding methods based on biased random walks that are guided according to known node labels. To design appropriate random walk strategies we start from the ``good-bad'' hubs perspective indicated by Radovanovi\'{c} et al.~\cite{radovanovic2010hubs}. The main idea is that the random walk prefers visiting good hubs and avoiding bad hubs. The proposed methods are evaluated by training widely used classification models on obtained embeddings. To demonstrate effectiveness of our methods we analyze the same classification models trained on node2vec embeddings. The node2vec algorithm is selected as the baseline since it is the most general-purpose graph embedding method based on biased random walks. The obtained results indicate that the designed random walking schemes significantly improve the performance of final classification models.

\section{Hub-based Random Walk Graph Embedding Methods}
\label{brwm}

Let $G$ be a  partially labeled graph,  i.e., in $G$ a certain fraction of nodes is labeled and we want to infer labels of unlabeled nodes. Let $L(x)$ denote the label of an arbitrary node $x$ (if $x$ is labeled). A high-degree  labeled node $h$ is called a good hub if
\begin{equation*}
\left| \{n \in N^{l}(h) \: : L(n) = L(h) \} \right| \: \: > \: \: \left| \{m \in N^{l}(h) \: : L(m) \neq L(h)\} \right|,
\end{equation*}
where $N^{l}(h)$ is the set of the labeled neighbours of $h$ and $\left|\cdot \right|$ is the set cardinality operator. If the previously stated condition is not satisfied then $h$ is a bad hub. In other words, a good hub shares the same label with a majority of nodes in its neighborhood. Such property guaranties that its label can be correctly derived from the most frequent label of surrounding nodes, which is not the case for bad hubs. 

In this Section we present two novel hub-based random walk graph embedding methods based on the principle of favoring good hubs and avoiding bad hubs during random walks. As in other similar methods, a certain number of random walks is sampled starting from each node in the graph and each random walk has a fixed length. A general form of random walk graph embedding methods is shown in Algorithm \ref{alg1}. Each random walk is treated as a sentence ($s$ in Algorithm \ref{alg1}) composed of identifiers of nodes visited during the walk. All sampled random walks then form a text (denoted by $T$ in Algorithm \ref{alg1}) which is substituted to a word embedding algorithm $W$. Our methods, similarly to other random walk graph embedding algorithms, use word2vec~\cite{Mikolov2013Ext} as the underlying word embedding algorithm. We assume that $G$ does not contain isolated nodes, so each node $c$ in $G$ has a non-empty neighborhood~\footnote{Isolated nodes in random walk graph embedding methods are typically handled by adding self-loops. In this way, a random walk sampled from an isolated node $o$ is a sentence composed entirely of $o$.} (denoted as $N^{c}$ in Algorithm \ref{alg1}). 

\begin{algorithm}[htb]
\label{alg1}
\small
\SetAlgoLined
\DontPrintSemicolon
\SetKwInOut{Input}{input}
\SetKwInOut{Output}{output}
\Input{$G = (V, E)$ -- input graph ($V$ -- the set of nodes, $E$ -- the set of edges)\\
$t$ -- the number of random walks per node\\
$l$ -- the length of each random walk\\
$S$ -- sampling strategy\\
$W$ -- word embedding algorithm
}
\Output{$\hat{E}$ -- an embedding of $G$} 

\BlankLine \BlankLine 

$T$ = [] \\
\For{$v \in V$} {
   \For{$i = 1$ {\bf to} $t$} {
      $s$ = [] \\
      $c$ = $v$ \\
      \For{$j = 1$ {\bf to} $l$} {
         append {\it id} of $c$ to $s$ \\
         $N^{c} = $ the set of neighbors of $c$ \\ 
         $n$ = select a node from $N^{c}$ according to $S$\\ 
         $c = n$
      }
      append $s$ to $T$
   }
}

$\hat{E}$ = apply $W$ to $T$\\
{\bf return} $\hat{E}$

\caption{{\bf General form of random walk graph embedding algorithms}}	
\SetAlgoRefName{alg1}
\SetAlgoCaptionSeparator{'.'}
\end{algorithm}

The most crucial part of random walk graph embedding algorithms is the sampling strategy ($S$ in Algorithm \ref{alg1}). $S$ determines the next node ($n$ in Algorithm \ref{alg1}) in the random walk considering the node at which the random walk is currently located (current node, denoted by $c$ in Algorithm \ref{alg1}) and eventually the node from which the random walk started (start node, denoted by $v$ in Algorithm \ref{alg1}). $S$ is an unbiased sampling strategy if each node from $N^{c}$ has an equal probability to be selected as the next node. An important property of the unbiased sampling strategy is that it actually favors selecting hubs: a high degree node has a higher probability to be in $N^{c}$ than a low degree node. This property actually inspired a class of mathematical models for generating networks with hubs that are also known as {\it copying} models~\cite{Kumar2000,Newman2003}. Consequently, to design a biased sampling strategy preferring visiting good hubs we should include a degree of uniformity when selecting the next node. 
In our methods, this is achieved in two ways:
\begin{enumerate}
\item with a certain probability $q$ we always select a node from $N^{c}$ uniformly at random, and 
\item the next node is selected from a subset of $N^{c}$ containing good nodes (nodes having the same label as the start node), either uniformly at random or by higher probabilities for good hubs.
\end{enumerate}
If the current node $c$ is unlabeled then it can not be determined if it is good or bad hub, thus the next node $n$ is selected uniformly at random from the set of $c$'s neighbors.

The probability $q$ actually controls to what extent our sampling strategies deviate from being unbiased (higher $q$ values imply more unbiased sampling).  The unbiased sampling is also applied if the subset of $N^{c}$ containing good nodes is an empty set. With $p$ we denote the probability of biased sampling  ($p = 1 - q$) which is one of parameters of our graph embedding methods.  

\begin{algorithm}[htb]
\label{alg2}
\small
\SetAlgoLined
\DontPrintSemicolon
\SetKwInOut{Input}{input}
\SetKwInOut{Output}{output}
\Input{$v$ -- start node \\
$c$ -- current node \\ 
$N^{c}$ -- neighbors of $c$ \\
$p$ -- the probability of biased sampling}
\Output{$n$ -- next node} 

\BlankLine \BlankLine
$r$ = random real number between 0 and 1\\
\uIf{$c$ and $v$ are labeled {\bf and} $r <= p$} {
	$S = \{ x \in N^{c} \: : \: L(x) = L(v) \}$ \\
	$D = \{ y \in N^{c} \: : \: L(y) \neq L(v) \}$ \\
	\uIf{$|S| > 0$} {
		$n$ = select a node from $S$ uniformly at random
	}\Else {
		$n$ = select a node from $D$ uniformly at random
	}
}\Else{$n$ = select a node from $N^{c}$ uniformly at random}

{\bf return} $n$

\caption{{\bf The sampling strategy for the \texttt{SCWalk} algorithm}}	
\SetAlgoRefName{alg2}
\SetAlgoCaptionSeparator{'.'}
\end{algorithm}

\begin{algorithm}[htb]
\label{alg3}
\small
\SetAlgoLined
\DontPrintSemicolon
\SetKwInOut{Input}{input}
\SetKwInOut{Output}{output}
\Input{$v$ -- start node \\
$c$ -- current node \\ 
$N^{c}$ -- neighbors of $c$ \\
$p$ -- the probability of biased sampling}
\Output{$n$ -- next node} 

\BlankLine \BlankLine
$r$ = random real number between 0 and 1\\
\uIf{$c$ and $v$ are labeled {\bf and} $r <= p$} {
    $H$ = a dictionary mapping nodes to their good hubness values \\
	\For{$x \in N^{c}$} {
	    $N^{x}$ = the set of neighbors of $x$ \\
	    $H[x] = |\{ y \in N^{x} \: : \: L(y) = L(v) \}| \: / |N^{x}|$\\
	}
	
	$P$ = normalize $H$ values to a probability distribution  \\
	$n$ = randomly select a node from $N_{c}$ according to $P$ \\
}\Else{$n$ = select a node from $N^{c}$ uniformly at random}
	
{\bf return} $n$
	
\caption{{\bf The sampling strategy for the \texttt{Hub\-Walk\-Distribution} algorithm}}	
\SetAlgoRefName{alg3}
\SetAlgoCaptionSeparator{'.'}
\end{algorithm}

Our first method is called \texttt{SCWalk} (same class walk). Its sampling strategy is shown in Algorithm \ref{alg2}. The main idea of \texttt{SCWalk} is to direct random walks towards nodes that have the same label as the start node. With the probability of biased sampling, the sampling strategy selects the next node uniformly at random from the set of good nodes ($S$ in Algorithm \ref{alg2}). If the set of good nodes is empty then the next node is selected uniformly at random from the set of bad nodes. In this way, sentences sampled for the start node dominantly contains good nodes (and good hubs due to uniform sampling) from its neighborhood, consequently making the corresponding label more compact in the produced embedding and with less noise caused by bad nodes (and bad hubs). 

Our second hub-based random walk graph embedding method is called \texttt{Hub\-Walk\-Distribution}. The sampling strategy of this method is given in Algorithm \ref{alg3}. Compared to \texttt{SCWalk}, \texttt{HubWalkDistribution} considers not only neighbors of the current node, but also their neighbors. Additionally, it puts more bias towards good hubs and at the same time penalizes bad hubs more. \texttt{HubWalkDistribution} utilizes a measure of good hubness to form the probability distribution for selecting the next node. More specifically, for each neighbor $x$ of current node $c$ we compute $H(x)$ reflecting the degree of good hubness of $x$. $H(x)$ is the number of neighbors of $x$ having the same label as the start node divided by the total number of $x$'s neighbors. Higher values of $H(x)$ imply good hubness, whereas lower value close to 0 indicate bad hubness. The next node is then selected according to the probability proportional to its $H$ value.


As can be seen from the given algorithmic descriptions of the proposed random walk sampling strategies: (1) random walks are sampled for both labeled and unlabeled nodes, (2) random walk transitions for unlabeled nodes are always unbiased (each neighbor of an unlabeled node has an equal probability to be selected as the next node in the walk) since their labels are unknown, and (3) unlabeled nodes are not discarded from biased random walks starting from labeled nodes (biased according to our label-based sampling strategies). Thus, we can distinguish between three types of random walks sampled according to our schemes:
\begin{itemize}
\item {\it completely-unbiased} random walks that are sampled starting from unlabeled nodes, 
\item {\it partially-biased} random walks that are sampled starting from labeled nodes and may contain unlabeled nodes, and
\item {\it completely-biased} random walks that are sampled starting from labeled nodes and contain only labeled nodes.
\end{itemize}
The main principle of graph embedding methods based on random walks is that all node embedding vectors are learnt from all sampled random walks by training exactly one language model (Word2Vec in our case) from which node embedding vectors are formed. Consequently, node embedding vectors of unlabeled nodes are affected not only by completely-unbiased random walks, but also by partially-biased random walks. The number of partially-biased random walks increases with the fraction of labeled nodes, producing a stronger impact on node embedding vectors of unlabeled nodes.

\section{Experiments and Results}
\label{experiments_results}

The experimental evaluation of our hub-aware graph embedding methods is conducted on datasets listed in Table~\ref{table_datasets}. The experimental set of graphs contains both undirected and directed graphs. Directed graphs are converted to their undirected projections (by ignoring link directions) since random walks should not be restricted and biased to out-neighborhoods of nodes (i.e., by allowing random walks to follow only out-going links we capture neighborhoods partially; additionally, if a directed graph is not strongly connected then random walks cannot be continued when they enter a node without out-going links).  

\begin{table}[htb]
	
	\begin{center}
		\begin{tabular}{llllllllll}
			\noalign{\smallskip}\hline \noalign{\smallskip}
			Graph & $N$ & $E$ & $C$  & $\bar{d}$ & std($d$) & max($d$) & $L$ \\
			\noalign{\smallskip}\hline \noalign{\smallskip}
			Zachary karate club & 34 & 78 & 1  & 4.59 & 3.88 & 17 & 2\\
			CORAML & 2995 & 8158 & 61 &  5.45 & 8.25 & 246 & 7\\
			CITESEER & 4230 & 5337 & 515 &  2.52 & 3.75 & 85 & 6\\
			AE photo & 7650 & 119081 & 136 & 31.13 & 47.28 & 1434 & 8\\
			PUBMED & 19717 & 44324 & 1 &  4.50 & 7.43 & 171 & 3\\
			CORA & 19793 & 63421 & 364 &  6.41 & 8.79 & 297 & 70\\
			DBLP & 17716 & 52867 & 589 &  5.97 & 9.35 & 339 & 4 \\
			\noalign{\smallskip}\hline \noalign{\smallskip} 
		\end{tabular}
		\caption{Experimental datasets.}
		\label{table_datasets}
	\end{center}
\end{table}

All examined graphs have entirely labeled nodes (where the number of distinct labels varies from 2 to 70, column $L$ in Table~\ref{table_datasets}). 
The experimental corpus contains one small social network (Zachary karate club), five medium to large citation networks (CORAML, CITESEER, PUBMED, CORA and DBLP) and one large co-purchase network of Amazon photo-related products (AE photo). The Zachary karate club network depicts social interactions among members of a university karate club that were documented and firstly investigated by anthropologist Wayne W. Zachary~\cite{Zac77}. The labels in this graph correspond correspond to two karate clubs formed after a conflict between two members of the original karate club. Thus, on this graph we analyze the performance of binary classification models. In all other graphs, the number of labels is higher than 2, which means that we train and analyze multi-class predictive models. The nodes in a citation network represent scientific papers, while links are citations among them. Names of citation networks indicate bibliographical databases from which those networks are formed. Labels in citation networks denote broad or more narrow scientific fields (depending on the network). Links in co-purchase networks connect products that were bought in same transactions, while labels represent product categories.

Besides the total number of labels, Table~\ref{table_datasets} shows for each graph the number of nodes ($N$), the number of edges ($E$), the number of connected components ($C$), the average node degree ($\bar{d}$), the standard deviation of node degrees (std($d$)) and the maximal degree (max($d$)). It can be noticed that graphs are sparse ($\bar{d} \ll N - 1$; $N - 1$ is the maximal number of links that could emanate from any node). Zachary karate club and PUBMED are connected graphs (there is a path connecting any pair of nodes). Other networks contains a relatively large number of connected components. However, except CITESEER, all of them have a giant connected component encompassing more than 90\% of nodes (the largest connected component in CITESEER contains approximately 40\% of nodes). It is also important to observe that in all graphs, except the smallest one, we have that $\mbox{std}(d) > \bar{d}$ implying that the degree distributions of networks are long-tailed. This means that all examined medium and large graphs from our experimental corpus contain hubs. This is also evident from the maximal degree values where we have that $\mbox{max}(d) \gg \bar{d}$. 

In the experimental evaluation we compute classification performance of three traditional classification models, support vector machines (SVM), na\"ive Bayes (NB) and random forests (RF) on embeddings obtained from node2\-vec and our two methods. For each dataset and each method we generate graph embeddings in the following five dimensions: 10, 25, 50, 100 and 200. Node2vec was tuned by finding values of its hyper-parameters $p$ (return-back parameter) and $q$ (in-out parameter) that give embeddings with the lowest graph reconstruction error by the procedure described in ~\cite{2021_NCLID_SISAP}. As suggested in~\cite{nodetovec2016}, for $p$ and $q$ of node2vec we consider values in \{0.25, 0.50, 1, 2, 4\}, while the number of random walks per node and the length of each random walk were fixed to 10 and 80, respectively. The same number of random walks per node and the length of each random walk are used for our methods. The probability of biased sampling ($p$ hyper-parametar of our two methods) is varied in \{0.15, 0.5, 0.85\}, which means that our experimental analysis covers three different cases: the first one is when our methods incline to unbiased random walks ($p = 0.15$), the second one in which unbiased sampling occurs at the same frequency as biased sampling ($p = 0.5$), and the last one is when biased sampling dominates over unbiased sampling ($p = 0.85$). 

Trained classification models are evaluated by the 10-fold cross-validation procedure~\cite{refaeilzadeh2009cross} in which we compute the macro-averaged values of precision, recall and $F_{1}$ scores from confusion matrices obtained in each cross-validation step. Since we use 10-fold cross-validation, in each cross-validation step labels of 10\% of nodes are ignored and inferred from the labels of the rest of the nodes. This means that in our experimental setting the probability that a node is unlabeled is equal to 0.1, which is lower than the probability $1 - p$ of unbiased sampling in all experimental cases (please recall that $p$ takes values in \{0.15, 0.5, 0.85\}). This ensures that label information in biased random walk sampling is used to an extent that does not overfit SVM, NB and RF classification models derived from sampled random walks. For example, when $p$ is equal to 0.85 (the case with the most strongest biased random walk sampling in our experiments), labels of 15\% of nodes are ignored by our graph embedding methods, while labels of 10\% of nodes are ignored during the cross-validation procedure for training and evaluating the classification models.

Let $C$ denote the set of classes (labels) and let $e$ be the confusion matrix obtained in an arbitrary cross-validation step, where $e_{i,j}$ is the number of test examples (nodes from the test dataset) belonging to class $i$ that were classified into class $j$. Then, the accuracy of a classification model is defined as the number of correctly classified test examples:
\begin{equation*}
\mbox{Accuracy} = \frac{\sum\limits_{i \in C}{e_{i,i}}}{\sum\limits_{(i, j) \in C \times C}{e_{i, j}}}.
\end{equation*}
Contrary to accuracy, precision and recall are computed per class and then averaged into single scores. The precision for class $i$ is the fraction of test examples assigned to class $i$ that actually belong to class $i$: 
\begin{equation*}
\mbox{Precision}(i) = \frac{e_{i,i}}{\sum\limits_{j \in C}{e_{j, i}}}.
\end{equation*}
Recall quantifies to what extent members of class $i$ have been classified correctly:
\begin{equation*}
\mbox{Recall}(i) = \frac{e_{i,i}}{\sum\limits_{j \in C}{e_{i, j}}}.
\end{equation*}
Since precision and recall reflects two different aspects of classification models it is useful to merge them into a single score reflecting the overall classification performance. For this purpose, $F_{1}$ score, defined as the harmonic mean of averaged precision and recall, is typically used:
\begin{equation*}
\mbox{F}_{1} = \frac{2 \cdot P \cdot R}{P + R}, 
\end{equation*}
where $P$ and $R$ denote the average precision and recall, respectively.

The rest of this section is divided into several subsections, where each subsection presents obtained experimental results for one of considered classification models.

\subsection{SVM Evaluation}

In this subsection we present evaluation results for SVM classifiers. The average values of classification performance metrics across all experimental datasets for SVMs trained on node2vec embeddings are shown in Table~\ref{avg_svm_n2v}. It can be observed that SVM on node2vec embeddings has a relatively good predictive power with all performance metrics being higher than 0.6. Table~\ref{avg_svm_walks} shows the average performance of SVM trained on embeddings produced by our methods for different values of the probability of biased sampling ($p$). It can be seen that regardless of $p$ our graph embedding methods outperform node2vec. Additionally, the predictive power increases with $p$ implying that a higher degree of hub-based biases in our sampling strategies gradually improves SVM classifiers.  

\begin{table}[htb!]
	
	\begin{center}
		\begin{tabular}{llll}
			\noalign{\smallskip}\hline \noalign{\smallskip} 
			\bf{Accuracy} & \bf{Precision} & \bf{Recall} & \bf{$F_{1}$}    \\
			\noalign{\smallskip}\hline \noalign{\smallskip} 
			0.7494 & 0.6802 & 0.6272 & 0.6419   \\
			\noalign{\smallskip}\hline \noalign{\smallskip}
		\end{tabular}
		\caption{Average values of performance metrics for SVM classification on node2vec embeddings.}
		\label{avg_svm_n2v}
	\end{center}
\end{table}

\begin{table}[htb!]
	
	\begin{center}
		\begin{tabular}{lllll}
			\noalign{\smallskip}\hline \noalign{\smallskip} 
			\bf{p} & \bf{Accuracy} & \bf{Precision} & \bf{Recall} & \bf{$F_{1}$}    \\
			\noalign{\smallskip}\hline \noalign{\smallskip}
			0.15 & 0.7677 & 0.6833 & 0.6378 & 0.6515   \\
			0.5 & 0.7919 & 0.7027 & 0.6587 & 0.6721   \\
			0.85 & 0.8148 & 0.7228 & 0.6804 & 0.6940   \\
			\noalign{\smallskip}\hline \noalign{\smallskip}
		\end{tabular}
		\caption{Average values of performance metrics for SVM classification on all embeddings generated by SCWalk and HubWalkDistribution.}
		\label{avg_svm_walks}
	\end{center}
\end{table}

Now, we are going to present performance improvements (or deterioration) when $p$ is equal to 0.85 (the case with the biggest average improvements over node2vec) for each graph and each of our graph embedding methods, individually. Let us denote differences in accuracy, precision, recall and $F_{1}$ by Acc-diff, Prec-diff, Rec-diff and F1-diff, respectively. Differences between SCWalk and node2vec are shown in Table~\ref{diff_sc} (a positive value indicates that SCWalk performs better). It can be seen that there are considerable improvements in performance metrics on almost all graphs (no negative values in Table~\ref{diff_sc}). SVM trained on SCWalk embeddings of the Zachary karate club graph has exactly the same values of performance metrics as SVM trained on node2vec embeddings. Due to a small set of nodes, SVM models on this graph achieve perfect performance without false predictions for both node2vec and our methods. For other graphs, accuracy, recall and $F_{1}$ of SCWalk is better compared to node2vec. Precision also strongly tend to be improved with SCWalk: we can seen positive differences in precision for all graphs except CITESEER where the difference is equal to zero. Significant improvements (improvements higher than 0.1) in all performance metrics are present for two graphs (CORA and CORAML).   

\begin{table}[htb!]
	
	\begin{center}
		\begin{tabular}{lllll}
			\noalign{\smallskip}\hline \noalign{\smallskip}  
			\bf{Graph} & \bf{Acc-diff} & \bf{Prec-diff} & \bf{Rec-diff} & \bf{$F_{1}$-diff}    \\
			\noalign{\smallskip}\hline \noalign{\smallskip} 
			AE photo & 0.0473 & 0.055 & 0.0488 & 0.052   \\
			CITESEER & 0.0572 & 0 & 0.0095 & 0.0079   \\
			CORA & 0.1462 & 0.1687 & 0.1571 & 0.1658   \\
			CORAML & 0.1016 & 0.1382 & 0.1313 & 0.1383   \\
			DBLP & 0.1863 & 0.0046 & 0.1152 & 0.0876   \\
			Zachary karate club & 0 & 0 & 0 & 0   \\
			PUBMED & 0.0867 & 0.0892 & 0.094 & 0.092   \\
			\noalign{\smallskip}\hline \noalign{\smallskip} 
		\end{tabular}
		\caption{Differences in SVM classification performance between SCWalk ($p=0.85$) and node2vec.}
		\label{diff_sc}
	\end{center}
\end{table}

\begin{figure}[htb!]
	\begin{center}
		\includegraphics[width=0.95\textwidth]{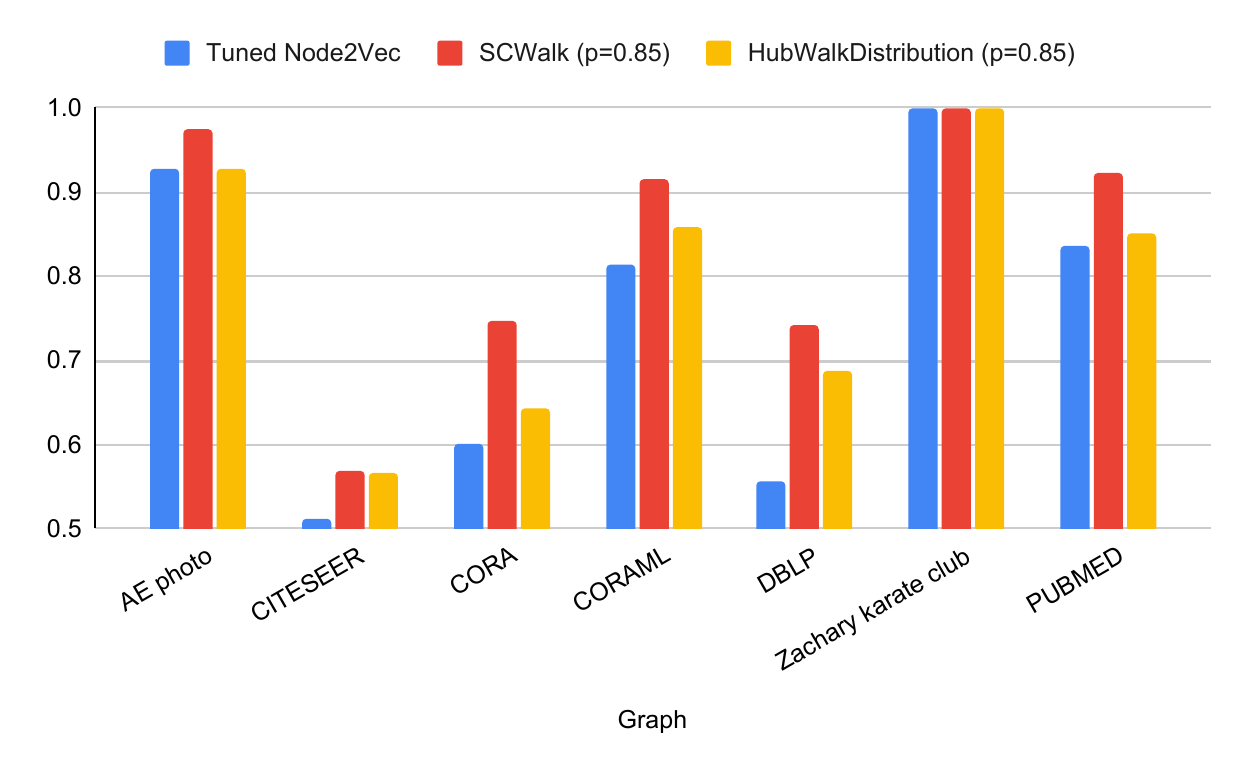}
	\end{center}
	\caption{Accuracy of SVM classification for all datasets and all three graph embedding algorithms.}
	\label{fig_avg}
\end{figure}

The differences between HubWalkDistribution and node2vec are shown in Table~\ref{diff_hub}. The improvements are present for all datasets except AE photo on which node2vec preforms slightly better than our algorithm. On the Zachary karate club graph we again obtain equal values of performance metrics. On other datasets, HubWalkDistribution achieves higher values of precision, recall and $F_{1}$ than node2vec, but with smaller improvements compared to  SCWalk. The accuracy of SVM models trained on embeddings produced by all three graph embedding methods for all datasets are shown in Figure~\ref{fig_avg}. It can be seen that SVM models trained after SCWalk considerably outperform SVM models trained after HubWalkDistribution and node2vec on medium and large graphs.     

\begin{table}[htb!]
	
	\begin{center}
		\begin{tabular}{lllll}
			\noalign{\smallskip}\hline \noalign{\smallskip}  
			\bf{Graph} & \bf{Acc-diff} & \bf{Prec-diff} & \bf{Rec-diff} & \bf{$F_{1}$-diff} \\
			\noalign{\smallskip}\hline \noalign{\smallskip} 
			AE photo & 0.0007 & -0.0013 & -0.0033 & -0.0023   \\
			CITESEER & 0.0539 & 0.0066 & 0.0129 & 0.01283 \\
			CORA & 0.0424 & 0.0598 & 0.0495 & 0.0553   \\
			CORAML & 0.0461 & 0.057 & 0.0476 & 0.0518   \\
			DBLP & 0.1321 & 0.0023 & 0.0668 & 0.0532   \\
			Zachary karate club & 0 & 0 & 0 & 0   \\
			PUBMED & 0.0144 & 0.0151 & 0.0151 & 0.0152   \\
			\noalign{\smallskip}\hline \noalign{\smallskip}  
		\end{tabular}
		\caption{Differences in SVM classification performance between HubWalkDistribution ($p=0.85$) and node2vec.}
		\label{diff_hub}
	\end{center}
\end{table}

\subsection{Random Forest Evaluation}

In the evaluation of RF classifiers we experiment with the number of estimators (the number of decision trees in a random forest model) to see how results depend on this hyper-parameter. The number of estimators is varied to be 10, 25, 50 and 100. The obtained results for node2vec are summarized in Table~\ref{rf_n2v}. As expected, the predictive performance of RF improves as the number of estimator grows, but not by a big margin. It can be also seen that RF on node2vec performs similarly to SVM on node2vec.

\begin{table}[htb!]
	
	\begin{center}
		\begin{tabular}{lllll}
			\noalign{\smallskip}\hline \noalign{\smallskip}   
			\bf{Estimators} & \bf{Accuracy} & \bf{Precision} & \bf{Recall} & \bf{$F_{1}$} \\
			\noalign{\smallskip}\hline \noalign{\smallskip}  
			10 & 0.7002 & 0.6757 & 0.6078 & 0.6219 \\
			25 & 0.7224 & 0.697 & 0.6259 & 0.6405   \\
			50 & 0.7286 & 0.7047 & 0.6308 & 0.6467   \\
			100 & 0.7315 & 0.7084 & 0.634 & 0.6495   \\
			\noalign{\smallskip}\hline \noalign{\smallskip}  
		\end{tabular}
		\caption{Average values of performance metrics for RF classification on node2vec embeddings.}
		\label{rf_n2v}
	\end{center}
\end{table}

The average values of performance metrics for RF models trained on SCWalk and HubWalkDistribution embeddings when $p=0.85$ are given in Table~\ref{rf_walk}. Again we can see that the predictive ability of RF increases with the number of estimators.
The obtained values of accuracy, precision, recall and $F_{1}$ of our graph embeddings methods are higher than the same values for node2vec. 
Consequently, our graph embedding algorithms are able to also improve the predictive power of RF classifiers.  

\begin{table}[htb!]
	
	\begin{center}
		\begin{tabular}{lllll}
			\noalign{\smallskip}\hline \noalign{\smallskip}
			\bf{Estimators} & \bf{Accuracy} & \bf{Precision} & \bf{Recall} & \bf{$F_{1}$} \\
			\noalign{\smallskip}\hline \noalign{\smallskip}
			10 & 0.7588 & 0.6922 & 0.6412 & 0.6544 \\
			25 & 0.7805 & 0.7106 & 0.6579 & 0.6735   \\
			50 & 0.7901 & 0.7181 & 0.6672 & 0.6825   \\
			100 & 0.7927 & 0.7215 & 0.6696 & 0.6857   \\
			\noalign{\smallskip}\hline \noalign{\smallskip}
		\end{tabular}		
		\caption{Average values of performance metrics for RF classification on all embeddings generated by SCWalk and HubWalkDistribution ($p=0.85$).}
		\label{rf_walk}
	\end{center}
\end{table}

\begin{figure}[htb!]
	\begin{center}
		\includegraphics[width=0.95\textwidth]{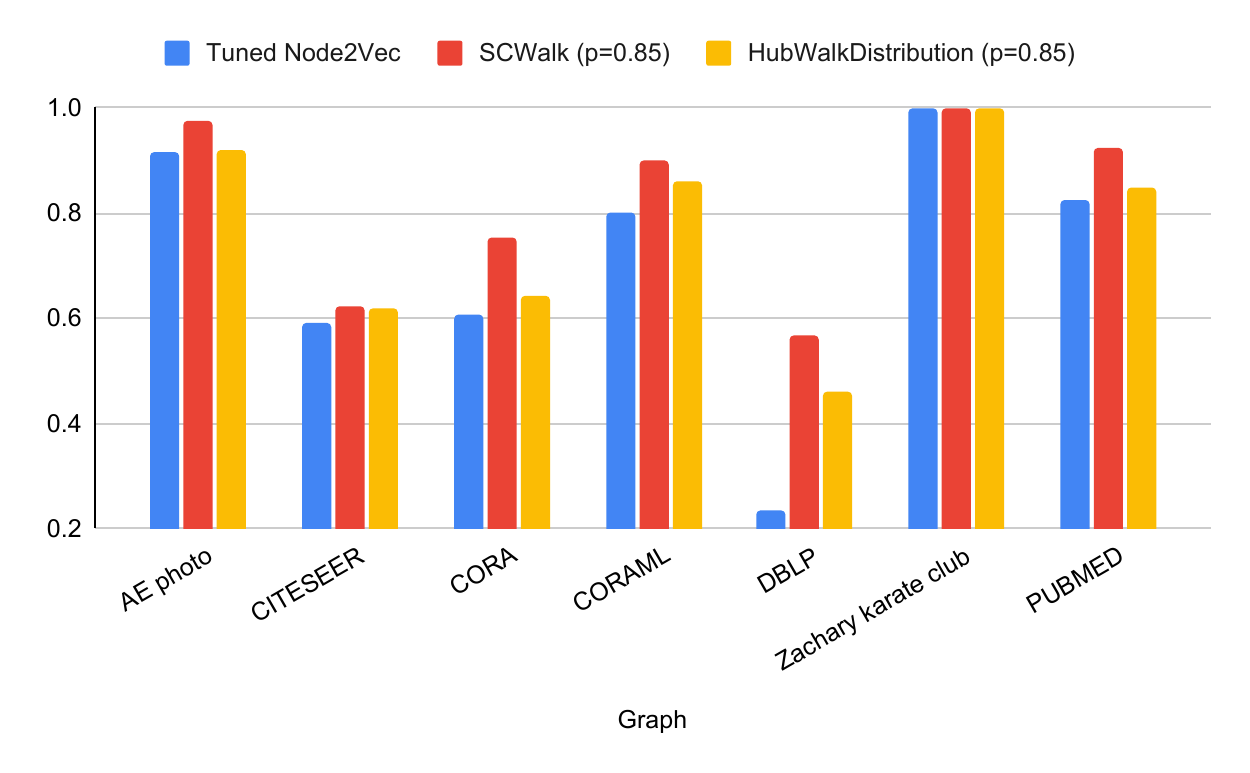}
	\end{center}
	\caption{Classification accuracy of RF with 100 estimators for each dataset and three graph embedding algorithms.}
	\label{fig_avg_rf}
\end{figure}

RF classifiers provide the best results on our datasets when the number of estimators is equal to 100. Figure~\ref{fig_avg_rf} shows a comparison of classification accuracy of RF for all three graph embedding methods on all graphs from our experimental corpus for the previously mentioned number of estimators. It can be seen that RF in combination with node2vec performs badly (accuracy just above 20\%) for DBLP. For that graph we see that our graph embedding methods significantly outperform node2vec with the increase of accuracy by almost 25\%. Except the Zachary karate club graph, RF trained on graph embeddings formed by our methods has a higher accuracy than RF trained on node2vec embeddings. The highest accuracy is achieved by SCWalk which significantly outperforms node2vec on 4 graphs (CORA, CORAML, DBLP and PUBMED).

\subsection{Na\"ive Bayes Evaluation}

In this subsection we present evaluation results for NB classification. The average values of performance metrics for node2vec and our random walk methods (for $p = 0.85$) are depicted in Tables \ref{avg_nb_n2v} and \ref{avg_nb_rw}, respectively. It can be observed that our methods on average provide better embeddings for NB classification than node2vec. 

\begin{table}[htb!]
	
	\begin{center}
		\begin{tabular}{llll}
			\noalign{\smallskip}\hline \noalign{\smallskip} 
			\bf{Accuracy} & \bf{Precision} & \bf{Recall} & \bf{$F_{1}$}    \\
			\noalign{\smallskip}\hline \noalign{\smallskip} 
			0.7034 & 0.6634 & 0.6304 & 0.634   \\
			\hline 
		\end{tabular}
		\caption{Average values of performance metrics for NB classification on node2vec embeddings.}
		\label{avg_nb_n2v}
	\end{center}
\end{table}

\begin{table}[htb!]
	
	\begin{center}
		\begin{tabular}{llll}
			\noalign{\smallskip}\hline \noalign{\smallskip} 
			\bf{Accuracy} & \bf{Precision} & \bf{Recall} & \bf{$F_{1}$}    \\
			\noalign{\smallskip}\hline \noalign{\smallskip}
			0.7553 & 0.6898 & 0.6496 & 0.6605   \\
			\noalign{\smallskip}\hline \noalign{\smallskip}
		\end{tabular}
		\caption{Average values of performance metrics for NB classification on all embeddings generated by SCWalk and HubWalkDistribution ($p=0.85$).}
		\label{avg_nb_rw}
	\end{center}
\end{table}

\begin{figure}[htb!]
	\begin{center}
		\includegraphics[width=0.95\textwidth]{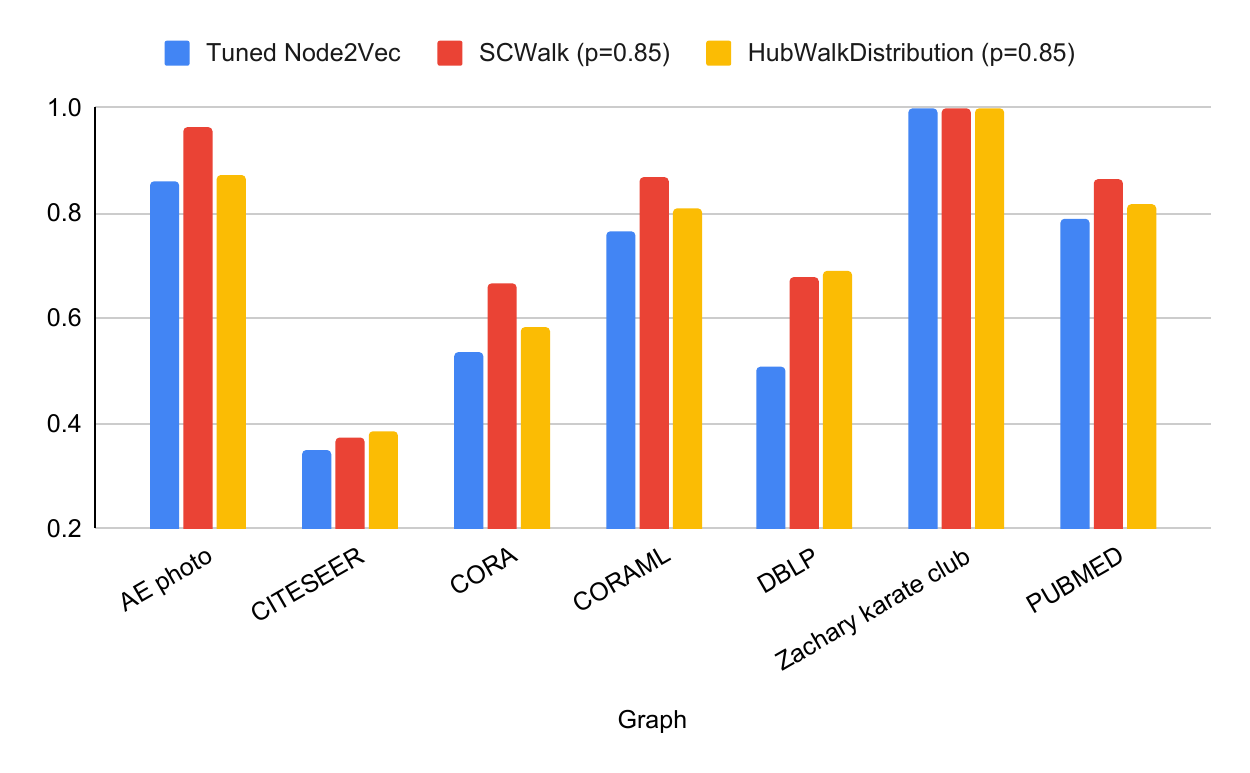}
	\end{center}
	\caption{Classification accuracy of NB for each dataset and three graph embedding algorithms.}
	\label{fig_avg_nb}
\end{figure}

Figure~\ref{fig_avg_nb} shows the classification accuracy of NB per dataset for three graph embedding algorithms. HubWalkDistribution is the best performing algorithm on 2 graphs (CITESEER and DBLP), while SCWalk is the best choice for 5 graphs (all other graphs except Zachary). It can be seen that SCWalk significantly outperforms node2vec for 5 graphs (AE photo, CORA, CORAML, DBLP and PUBMED). 

\subsection{Experimental Analysis of Hyper-parameters}

The principal difference of our graph embedding methods compared to existing graph embedding approaches based on biased random walks is leveraging node labels according to the previously described hubness principle. Thus, the most important hyper-parameter of our methods is the probability $p$ of biased random walk sampling. Larger values of $p$ imply that node labels are more frequently used in the sampling process. Consequently, for an increase in $p$ we also expect an increase in the accuracy of node classification models. In this subsection, we discuss in detail the impact of $p$ on the performance of classification models trained on SCWalk and HubWalkDistribution embeddings. 

Table~\ref{svm_results} shows accuracy, precision, recall and $F_{1}$ of SVM classification models trained on SCWalk and HubWalkDistribution embeddings obtained for different values of hyper-parameter $p$. The previously mentioned model evaluation metrics are averaged across all examined embedding dimensions (10, 25, 50, 100 and 200). It can be seen that with larger $p$ values we obtain better-performing SVM classifiers, i.e., accuracy, precision, recall and $F_{1}$ score of SVM models consistently increase with $p$. Considerable improvements can be observed for SCWalk where accuracy, precision, recall and $F_{1}$ increased by 8.92\%, 9.07\%, 10.58\% and 10.27\%, respectively, when $p$ goes from 0.15 to 0.85. Additionally, SVM classifiers trained on SCWalk embeddings provide better classification results than the same classifiers trained on HubWalkDistribution embeddings.

\begin{table}[htb!]
\begin{center}
\begin{tabular}{lllllllll}
\noalign{\smallskip}\hline \noalign{\smallskip} 
 & \multicolumn{4}{l}{{\bf SCWalk}}  & \multicolumn{4}{l}{{\bf HubWalkDistribution}} \\
\noalign{\smallskip}\hline \noalign{\smallskip}
\bf{$p$} & \bf{$A$} & \bf{$P$} & \bf{$R$} & \bf{$F_{1}$} & \bf{$A$} & \bf{$P$} & \bf{$R$} & \bf{$F_{1}$}  \\
\noalign{\smallskip}\hline \noalign{\smallskip}
0.15 & 0.7701 & 0.6834 & 0.6391 & 0.6526 & 0.7654 & 0.6831 & 0.6366 & 0.6504  \\
0.5 & 0.8085 & 0.7171 & 0.6743 & 0.6884 & 0.7752 & 0.6882 & 0.6431 & 0.6558   \\
0.85 & 0.8388 & 0.7454 & 0.7067 & 0.7196 & 0.7908 & 0.7002 & 0.6542 & 0.6684  \\
\noalign{\smallskip}\hline \noalign{\smallskip}
\end{tabular}
\caption{The average values of performance metrics for SVM classification on SCWalk and HubWalkDistribution embeddings for different values of $p$. $A$ -- accuracy, $P$ -- precision, $R$ -- recall}
\label{svm_results}
\end{center}
\end{table}

The results for RF and NB classification models are summarized in Tables~\ref{rf_results} and~\ref{nb_results}. Again, it can be seen a consistent increase in model performance for higher $p$ values for both SCWalk and HubWalkDistribution. As for SVM, RF and NB models trained on SCWalk embeddings are better than models trained on HubWalkDistribution embeddings. For SCWalk, the $F_{1}$ score of RF and NB models increased by 12.44\% and 11\%, respectively, when $p$ increases from 0.15 to 0.85.

\begin{table}[htb!]
\begin{center}
\begin{tabular}{lllllllll}
\noalign{\smallskip}\hline \noalign{\smallskip} 
 & \multicolumn{4}{l}{{\bf SCWalk}}  & \multicolumn{4}{l}{{\bf HubWalkDistribution}} \\
\noalign{\smallskip}\hline \noalign{\smallskip}
\bf{$p$} & \bf{$A$} & \bf{$P$} & \bf{$R$} & \bf{$F_{1}$} & \bf{$A$} & \bf{$P$} & \bf{$R$} & \bf{$F_{1}$}  \\
\noalign{\smallskip}\hline \noalign{\smallskip}
0.15 & 0.7328 & 0.6841 & 0.6184 & 0.6348 & 0.7253 & 0.6783 & 0.6112 & 0.6273  \\
0.5 & 0.7782 & 0.7124 & 0.6582 & 0.6739  & 0.744 & 0.6878 & 0.6236 & 0.6409  \\
0.85 & 0.8203 & 0.7448 & 0.6984 & 0.7138 & 0.7651 & 0.6982 & 0.6408 & 0.6577  \\
\noalign{\smallskip}\hline \noalign{\smallskip}
\end{tabular}
\caption{The average values of performance metrics for RF classification on SCWalk and HubWalkDistribution embeddings for different values of $p$. $A$ -- accuracy, $P$ -- precision, $R$ -- recall}
\label{rf_results}
\end{center}
\end{table}

\begin{table}[htb!]
\begin{center}
\begin{tabular}{lllllllll}
\noalign{\smallskip}\hline \noalign{\smallskip} 
 & \multicolumn{4}{l}{{\bf SCWalk}}  & \multicolumn{4}{l}{{\bf HubWalkDistribution}} \\
\noalign{\smallskip}\hline \noalign{\smallskip}
\bf{$p$} & \bf{$A$} & \bf{$P$} & \bf{$R$} & \bf{$F_{1}$} & \bf{$A$} & \bf{$P$} & \bf{$R$} & \bf{$F_{1}$}  \\
\noalign{\smallskip}\hline \noalign{\smallskip}
0.15 & 0.7085 & 0.6419 & 0.6104 & 0.6146 & 0.7054 & 0.6389 & 0.6079 & 0.612  \\
0.5 & 0.7421 & 0.6732 & 0.6367 & 0.6457 & 0.7214 & 0.6493 & 0.6181 & 0.6232  \\
0.85 & 0.7737 & 0.7138 & 0.668 & 0.6823 & 0.7368 & 0.6659 & 0.6311 & 0.6387  \\
\noalign{\smallskip}\hline \noalign{\smallskip}
\end{tabular}
\caption{The average values of performance metrics for NB classification on SCWalk and HubWalkDistribution embeddings for different values of $p$. $A$ -- accuracy, $P$ -- precision, $R$ -- recall}
\label{nb_results}
\end{center}
\end{table}

\section{Conclusions and Future Work}

Node classification is one of the most important machine learning task when analyzing partially labeled networks. In this paper we have discussed three different approaches to the node classification problem: collective inference methods, node classification based on graph embeddings and graph neural networks. The approach based on graph embeddings provides a valuable trade-off between classification accuracy and computational efficiency. However, graph embedding algorithms are typically designed to be independent of a concrete application, i.e., they provide general-purpose representations of nodes in Euclidean spaces that are suitable for a variety of tasks (including also node classification). 

In this paper we have presented two novel graph embedding algorithms (SCWalk and HubWalkDistribution) based on random walks that are specifically tailored for the node classification problem. This means that our algorithms explicitly take into account existing labels of nodes when sampling random walks. Additionally, our algorithms give a special attention to hubs, which are the most important nodes in large-scale networks. 

In the experimental evaluation, we have compared our methods to node2vec, which is the most popular general-purpose graph embedding algorithm based on random walks. More specifically, we have examined predictive performance of four traditional classification algorithms (SVM, RF and NB) trained on embeddings produced by our methods and node2vec on seven real-world networks (one small and six medium to large). The obtained results show that the predictive power of classification models improves with our methods. Additionally, it was demonstrated that a higher degree of hub-based biases when sampling random walks leads to more suitable embeddings for the node classification problem. 

Our future work can go in few directions. One of them will be to consider even more biased random walks approaches for graph embedding algorithms, e.g., to take node degree (or other centrality metrics) when computing transition probabilities during random walks. The set of target classification algorithms could be also extended when experimentally evaluating our methods, e.g., by including neural networks. The robustness of random walk sampling strategies to adversarial attacks is also an important issue. Therefore, it is valuable to test how our methods behave when labels are randomly changed for a small number of nodes. It is also interesting to examine SCWalk and HubWalkDistribution algorithms in other machine learning applications, such as link prediction and node clustering. 

\section*{Acknowledgements}

\noindent
This research was supported by the Science Fund of the Republic of Serbia, \#6518241, Graphs in Space: Graph Embeddings for Machine Learning on Complex Data -- GRASP, and \#7462, Graphs in Space and Time: Graph Embeddings for Machine Learning in Complex Dynamical Systems -- TIGRA.

\bibliography{refs}

\end{document}